\documentclass[10pt]{article} 
\usepackage[preprint]{tmlr}

\usepackage{amsmath,amsfonts,bm}

\def\eqref#1{equation~\ref{#1}}

\def\1{\bm{1}}

\DeclareMathAlphabet{\mathsfit}{\encodingdefault}{\sfdefault}{m}{sl}
\SetMathAlphabet{\mathsfit}{bold}{\encodingdefault}{\sfdefault}{bx}{n}

\newcommand{\selecteddata}{\DataAux'}

\newcommand{\DataTarget}{S_{\mathrm{tg}}}
\newcommand{\DataSource}{S_{\mathrm{src}}}
\newcommand{\DataAux}{S_{\mathrm{aux}}}

\newcommand{\forgetset}{S_{\mathrm{non-static}}}
\newcommand{\retainset}{S_{\mathrm{static}}}

\newcommand{\ModelSource}{w_{\mathrm{src}}}
\newcommand{\ModelTarget}{w_{\mathrm{tg}}}
\newcommand{\ModelAny}{w}

\newcommand\optparen[1]{\ifthenelse{\equal{#1}{}}{}{(#1)}}

\newcommand{\Alg}{\mathcal{A}}
\newcommand{\Unlearn}{\mathcal{U}}

\usepackage{url}

\usepackage{microtype}
\usepackage{graphicx}
\usepackage{caption}
\usepackage{subcaption}
\usepackage{booktabs} 

\usepackage{hyperref}

\usepackage{xparse}
\usepackage{xstring}
\usepackage{xspace}

\usepackage{amsmath}
\usepackage{amssymb}
\usepackage{mathtools}
\usepackage{amsthm}

\usepackage[capitalize,noabbrev]{cleveref}

\theoremstyle{plain}
\newtheorem{theorem}{Theorem}[section]

\theoremstyle{definition}
\newtheorem{definition}[theorem]{Definition}

\theoremstyle{remark}

\usepackage[textsize=tiny]{todonotes}

\newcommand{\defn}[1]{\emph{#1}}

\newcommand{\zerodel}{.\kern-\nulldelimiterspace}

\usepackage[hide=true,marginparwidth=1in]{marginalia}

\title{Data Selection for Transfer Unlearning}

\author{\name Nazanin Mohammadi Sepahvand \email sepahvan@mila.quebec \\
      \addr Electrical and Computer Engineering Department\\
      McGill University, Mila
      \AND 
      \name Vincent Dumoulin \email vdumoulin@google.com\\
      \addr Google DeepMind
      \AND 
      \name Eleni Triantafillou$^*$ \email etriantafillou@google.com\\
      \addr Google DeepMind
      \AND
      \name Gintare Karolina Dziugaite$^*$ \email gkdz@google.com\\
      \addr Google DeepMind \\
      McGill University, Mila
      }

\def\month{MM}  
\def\year{YYYY} 
\def\openreview{\url{https://openreview.net/forum?id=XXXX}} 

\begin{document}

\maketitle

\def\thefootnote{*}\footnotetext{Equal contribution.}

\begin{abstract}
As deep learning models are becoming larger and data-hungrier, there are growing ethical, legal and technical concerns over use of data: in practice, agreements on data use may change over time, rendering previously-used training data impermissible for training purposes. These issues have driven increased attention to machine unlearning: removing ``the influence of'' a subset of training data from a trained model. In this work, we advocate for a relaxed definition of unlearning that does not address privacy applications but targets a scenario where a data owner withdraws permission of use of their data for training purposes. In this context, we consider the important problem of \emph{transfer unlearning} where a pretrained model is transferred to a target dataset that contains some ``non-static'' data that may need to be unlearned in the future. 
    We propose a new method that uses a mechanism for selecting relevant examples from an auxiliary ``static'' dataset, and finetunes on the selected data instead of ``non-static'' target data; addressing all unlearning requests ahead of time. We also adapt a recent relaxed definition of unlearning to our problem setting and demonstrate that our approach is an exact transfer unlearner according to it, while being highly efficient (amortized).
    We find that our method outperforms the gold standard ``exact unlearning'' (finetuning on only the ``static'' portion of the target dataset) on several datasets, especially for small ``static'' sets, sometimes approaching an upper bound for test accuracy. We also analyze factors influencing the accuracy boost obtained by data selection.
\end{abstract}

\section{Introduction}
In recent years, deep learning systems have made tremendous progress in producing more general-purpose models that are able to tackle a wide range of downstream tasks \citep{zhou2023comprehensive}. A key ingredient in building modern deep learning pipelines is transfer learning: reusing a model trained on one or more \textit{source} task(s) in order to tackle a different \textit{target} task. This methodology is ubiquitous and increasingly useful in the era of large general-purpose pre-trained models: it enables effectively amortizing the cost of solving several tasks (the pretraining phase is very expensive while transfer learning can be significantly cheaper), saving both compute and energy, as well as facilitating stronger performance on target tasks that have limited training data available (where training a new model from scratch may lead to overfitting). 

Meanwhile, the prominent data-dependence of deep learning models has led to growing ethical, legal and technical concerns. Practically, for a variety of reasons, the permission to use data for different purposes may change over time. This poses technical challenges for deep learning pipelines: given a deep model that has already been trained on data that is no longer permissible, one must design strategies to efficiently modify it to ``remove the effect'' of the no-longer-permissible data, without unnecessarily degrading the utility of the model. This is challenging in highly non-convex networks where (efficiently) tracing the influence of data on trained weights is an open problem. Machine unlearning is a young but rapidly-growing research field that aims to achieve this goal; see \citet{nguyen2022survey} for a survey. 

Unlearning work is typically motivated by regulations like EU’s General Data Protection Regulation \citep{mantelero2013eu} that stipulate that individuals have the ``right to be forgotten'' and can request to have their data deleted. However, we argue that protecting user privacy is not the only use case of unlearning. Furthermore, both the definition as well as the evaluation metrics should be tailored to the application of interest, and we argue that previous proposals that are designed for privacy may be too stringent for other use cases. Instead, we consider a relaxed formulation of unlearning for non-privacy applications. A data owner may reconsider their decision to allow the direct use of their data to train models, but they may also want to keep benefiting from the representativity that comes with being part of the development cycle. In this hypothetical scenario, the data owner may modify their agreement to forbid the direct use of their data to train models but authorize indirect uses, such as for validation purposes.

In this work, we study \emph{transfer unlearning} for non-privacy applications: the problem of transferring a pretrained model to a target dataset that contains ``non-static'' data. In particular, our assumption is that pretraining occurred on ``static data'' that is not subject to modifications; an assumption that can be met easily in many practical cases. On the other hand, the target task contains some data that is ``non-static'' in the sense that their permission for use for training may be withdrawn. We therefore desire a transfer learning approach that leads to good performance on the target task while efficiently supporting unlearning requests.

To address this problem, one could simply finetune the pretrained model on all of the target data and then handle unlearning requests of target data as they come. 
In principle, this could be achieved by applying known unlearning algorithms that either guarantee the removal of the influence of the specified examples, e.g. via retraining (parts of) the model, or approximate this goal by running an unlearning procedure that  \textit{reduces} that influence \citep{golatkar2020eternal,goel2022towards,kurmanji2023towards}. However, both of these options have drawbacks. Retraining the model each time that a deletion request is issued is computationally expensive: while clever ensemble-based techniques can improve efficiency in some cases \citep{bourtoule2021machine,yan2022arcane}, they are comparable to naive retraining in the worst case. Further, ``approximate unlearning'' approaches either don't scale well or lack theoretical guarantees, making it hard to quantify the extent to which they achieve unlearning and necessitating careful (expensive) evaluation.

Therefore, we propose an alternative avenue: we design methodology that allows us to select the most ``relevant'' data points from an \textit{auxiliary dataset}, such that finetuning the pretrained model on those points instead of finetuning it on the non-static target data will yield good performance on the target task. Notably, we only use the non-static target data for selection purposes, but we never finetune on that data. We therefore effectively handle all deletion requests automatically ahead of time, eliminating the need for expensive or approximate unlearning as postprocessing. 

While this approach doesn't satisfy a common definition of exact unlearning, we argue that it is useful in practical scenarios to consider a different, relaxed, framework for unlearning. To that end, we adapt a recently proposed relaxed notion of unlearning to accommodate for data-dependent training data selection mechanisms for transfer unlearning. We demonstrate that our proposed approach is an exact unlearner according to that definition, while being more efficient than alternatives: it can be viewed as amortizing all future unlearning requests from the non-static portion of the data. Our thorough empirical investigation reveals that our data selection method is effective and narrows the gap in terms of target test accuracy to an upper bound obtained by training on non-static data without unlearning. We further conduct analyses to understand when data selection is most effective.
Finally, we note that our approach is also robust against attacks that try to extract training data by comparing the model pre- and post-unlearning: our model does not change when new unlearning requests come in, while standard unlearning approaches, including retraining from scratch, may expose the information about the removed training data.
\ \\

{\bf Contributions.}
\begin{itemize}
    \item We formalize and study transfer unlearning and we  propose a novel method that replaces non-static target data by selected examples from an auxiliary static dataset, addressing all possible unlearning requests in an amortized way;
    \item We adapt a recently-proposed relaxed definition of unlearning to support the use of data selection in the context of transfer learning and we demonstrate that our approach is an exact unlearner according to it;
    \item We conduct a thorough empirical investigation and find that our method can, in many datasets, outperform the \NA{gold standard}\fTBD{GKD: not in love with this but cannot think of another way that would communicate that it is a reasonable baseline} baseline that finetunes the model on only the static portion of the target data, especially when the size of that portion is small;
    \item In some cases, our method can even bridge the gap (in target test accuracy) to the upper bound achieved by finetuning on all target data without unlearning;
    \item We analyze when our method is most effective, and highlight ``domain affinity'' between the auxiliary and target datasets as an important factor.
\end{itemize}

\section{Background}
\subsection{Unlearning Definitions}\label{sec:unlearningdefs}

Given a trained model $\Alg(S)$ and 
an arbitrary subset of training data $S' \subseteq S$ 
the goal of an unlearning algorithm $\Unlearn$ is to ``delete the influence'' of the data $S'$ from $\Alg(S)$. 
In this work, following previous literature \citep{golatkar2021mixed}, we assume that the training data can be partitioned into two subsets $S = \retainset \cup \forgetset$, where  ``static'' data is not subject to modifications whereas the permissions surrounding the use of ``non-static'' data may change. We study the case in particular where parts of the ``non-static'' data that was previously permissible to train on is no longer permissible to be used in for training, therefore necessitating unlearning of such data from any models that trained on them.

The following definition (a generalization of a definition from \citealt{ginart2019making}) formalizes this idea, in a manner reminiscent of guarantees from differential privacy \citep{dwork2006differential}. In order to state the definition concisely, we say two distributions $\mu,\nu$ are \defn{$(\epsilon,\delta)$-close} if 
$\mu(B) \le e^\epsilon \nu(B) + \delta$
and $\nu(B) \le e^\epsilon \mu(B) + \delta$ for all measurable events $B$. (This notion of divergence can be replaced by others to obtain related notions of unlearning.)

\begin{definition}
\label{defn:dpunlearning}
\textbf{Unlearning.} An unlearning algorithm $\Unlearn$ is an \defn{$(\epsilon,\delta)$-unlearner} (for $\Alg$) if, for all datasets $S = \retainset \cup \forgetset$ and subsets $S' \subseteq \forgetset$,
the distribution of 
$\Alg(S \setminus S')$ and  
$\Unlearn (\Alg(S),S')$ are $(\epsilon,\delta)$-close.
\end{definition}

The subset $S'$ is referred to in the literature as the ``forget set''. In this work, as described above, we assume that the forget set may only originate from the ``non-static'' portion of the dataset. Many variations of this definition exist. For example, 
$\Unlearn (\Alg(S),S',T(S))$ provides the unlearning algorithm 
also with some statistics of the original training data, which is known to be necessary for optimal rates \citep{sekhari2021remember}. 
(The same authors also advocate for other minor changes on technical grounds, but we will not dwell on these technical matters here.)

While this definition seems to be sufficient from the perspective of protecting sensitive data, verifying whether an algorithm meets this definition can be challenging, as it requires to compare to a model obtained without any influence of $S'$ on the optimization.
For example, if a training set $S$ is used both to train a model as well as make other decisions about the training algorithm $\Alg$ (e.g. optimization decisions, hyperparameter choices, training data subset choices), then to unlearn a subset $S' \subset S$, applying this same $\Alg$ on $S \setminus S'$ is not sufficient, since this doesn't remove the impact that $S'$ may have had on instantiating $\Alg$ itself. 
\citet{Anonymous} thus argue for the need for a relaxed definition and propose the following notion of unlearning \emph{relative to a class of learning algorithms}, rather than to a single algorithm.
\begin{definition}
\label{defn:unlearninggeneral}
\textbf{Relative Unlearning.} Consider a parameterized family $\{\Alg_\theta\}_{\theta \in \Theta}$ of learning algorithms and a (possibly randomized) map $S \mapsto \hat\theta(S)$ from datasets into parameters.
An unlearning algorithm $\Unlearn$ is an \defn{$(\epsilon,\delta)$-unlearner (relative to $\{\Alg_\theta\}_{\theta \in \Theta}$ and $\hat\theta$)} if, 
for all datasets $S = \retainset \cup \forgetset$ and subsets $S' \subseteq \forgetset$,
given $\hat\theta(S)$, the conditional distributions of 
$\Alg_{\hat\theta(S)}(S \setminus S')$ and  
$\Unlearn (\Alg_{\hat\theta(S)}(S),S')$ are $(\epsilon,\delta)$-close.
\end{definition}

The choice of the map $\hat\theta$ dictates how stringent this notion of unlearning is.

\newcommand{\defeq}{\overset{\mathrm{def}}{=\joinrel=}}

\subsection{Transfer Learning}\label{sec:transferlearning}

Transfer learning refers to the problem setting where a model that is trained on one or more \textit{source} tasks is adapted to tackle a different \textit{target} task.  In this work, we consider a realistic scenario, where we have \emph{target} data, $\DataTarget$, and a pretrained source model, $\ModelSource$, but do not necessarily have access to the \emph{source} (training) data,  $\DataSource$, or training recipe.

From now on, we will use $S$ to denote the available training data of the transfer learning algorithm. Commonly $S = \DataTarget$, but we will later consider a case where $S = \DataTarget \cup \DataAux$ also contains an auxiliary dataset $\DataAux$.

We will write $\ModelTarget = \Alg(\ModelSource, S)$ for the \emph{transfer learning algorithm $\Alg$}, acting on the pretrained model and training data $S$. Note that $\Alg(\ModelSource, \cdot) $ is an ordinary learning algorithm.


\subsection{Transfer Unlearning}\label{sec:transferunlearning}

Departing from the standard formulation of transfer learning, we study an important generalization of the problem, where some subset of the target data, $\DataTarget$, may be requested to be deleted. 
\emph{Transfer unlearning} is the problem of transferring a source model, that was trained on a static source dataset, to a target task, in a way that yields good accuracy on (held-out data from) that task while also supporting deletion requests of non-static target data efficiently.

As before, we assume that the target can be written $\DataTarget = \retainset \cup \forgetset$ and the training data for transfer learning is $S = \DataTarget \cup \DataAux =\retainset \cup \forgetset \cup \DataAux$ with potentially non-empty $\DataAux$.
In our experiments, we vary the ratio of static to non-static data, including the extreme case where the entire target dataset is non-static and therefore subject to deletion requests.

In this unlearning setting, 
given a pretrained model $\ModelSource$,
the goal is to produce two algorithms: 
a transfer learning algorithm $\Alg: (\ModelSource, S) \mapsto \ModelTarget$ and an unlearning algorithm $\Unlearn$.
That is, we wish to be able to produce an adapted model $\ModelTarget$ that works well on the target task, while having the ability to service deletion requests from $\forgetset$ efficiently via an unlearning algorithm.

\begin{definition}
\label{defn:dpunlearning-transfer}
\textbf{Transfer Unlearning.}
An algorithm $\Unlearn$ is an \defn{$(\epsilon,\delta)$-transfer-unlearner} (for $\Alg$) if $\Unlearn$ is an
$(\epsilon,\delta)$-unlearner for $\Alg(\ModelSource,\cdot)$, for all pretrained models $\ModelSource$.
\end{definition}

\begin{definition}
\label{defn:unlearninggeneral-transfer}
\textbf{Relative Transfer Unlearning.}
An unlearning algorithm $\Unlearn$ is an \defn{$(\epsilon,\delta)$-transfer-unlearner (relative to $\{\Alg_\theta\}_{\theta \in \Theta}$ and $\hat\theta$)}
if $\Unlearn$ is 
an $(\epsilon,\delta)$-unlearner relative to $\{\Alg_\theta(\ModelSource,\cdot)\}_{\theta \in \Theta}$ and $\hat\theta(\ModelSource,\cdot)$, for all models $\ModelSource$.
\end{definition}

Later, we will propose a method that assumes access to an auxiliary static dataset (an assumption that is easily met in several practical scenarios) and utilizes a data selection mechanism to select the most relevant auxiliary data from that dataset to replace non-static target examples. We will demonstrate that it satisfies relative transfer unlearning.

\section{Related Work}
\label{sec:relatedwork}

Our work lies at the intersection of transfer learning, unlearning, and data selection, which we review in this section.

{\bf Transfer learning.} Transfer learning aims to design adaptation methods for effective and efficient adapation to the target task. 
Various approaches to adapting the pretrained network exist, and linear probing (training a new classification layer on the frozen representation) or finetuning the entire network are perhaps the most ubiquitous. Alternatives include training a classification layer on a selection of intermediate representations in the network \citep{evci2022head2toe}, selective finetuning \citep{guo2019spottune,fu2021learn}, and training additional network modules (adapters) on top of the frozen network \citep{rebuffi2017learning,houlsby2019parameter,puigcerver2020scalable}.


{\bf Machine unlearning.} Unlearning, first coined in \citet{cao2015towards}, is the problem of removing from a trained network the influence of a subset of its training dataset. Unlearning is recently enjoying increased attention from the community \citep{neurips-2023-machine-unlearning}, driven by the promise to tackle important applications like removing biases from models, resolving confusion due to mislabelled training data, protecting user privacy, reducing hallucinations, improving forward transfer, to name a few \citep{goel2022towards,kurmanji2023towards,yao2023large,pawelczyk2023context}.

Unlearning in deep models is very challenging: these models are highly non-convex, making it hard to accurately trace the influence of data points on trained weights and other optimization choices, which in turn makes it difficult to efficiently surgically remove the influence of only the specified forget set, without harming the model's utility (e.g. accuracy on retained and held-out data). 

Several unlearning approaches have been proposed and are commonly categorized into \textit{exact} and \textit{approximate}, according to Definition \ref{defn:dpunlearning}. 
For deep neural networks, exact unlearning takes the form of retraining from scratch excluding the requested data. Ensemble-based techniques have been proposed for more efficient retraining (only the affected ensemble members) \citep{bourtoule2021machine,yan2022arcane}. 
However, in the worst case where forgetting requests are distributed over all ensemble members, these methods are similarly inefficient as naive retraining. On the other hand, approximate unlearning can generally be substantially more efficient, at the cost of potentially not fully eliminating the influence of requested examples \citep{golatkar2020eternal,golatkar2020forgetting,graves2021amnesiac,goel2022towards,thudi2022unrolling}. 

Most similar to our work, \citet{golatkar2021mixed} study unlearning in downstream tasks, using a pretrained model (that was trained on static data). However, they focus on approximate unlearning approaches while we instead argue for a relaxed notion of unlearning that is valuable in practical scenarios and demonstrate that our approach that utilizes data selection mechanisms offers exact unlearning under that relaxed notion while being efficient and outperforming alternative approaches.



{\bf Data selection.} \citet{chen2021weighted} propose a framework to learn weights over a set of source tasks for the purpose of improving downstream target task performance. 
They achieve this via a procedure that alternates between optimizing the source losses and optimizing the source task weights. 
Unfortunately, this method requires knowledge of the target task in advance. In contrast, we consider simpler schemes that don't require prior knowledge of transfer tasks nor tailored training algorithms and can instead utilize any off-the-shelf pretrained model.
Alternative data selection approaches have also been employed for boosting out-of-distribution performance and model robustness without a particular downstream task shift in mind \citep{yang2023identifying}.

\citet{wong2022dataset} proposed a ``dataset projection'' approach that finds the most related subset of an auxiliary dataset to co-train with a target dataset. Similar to \citet{chen2021weighted}, their approach also requires prior knowledge of (and a specialized training phase for) the target task.

\citet{liu2021improved} is the most similar to our work: they also aim to leverage pretrained models and apply the co-training with cleverly chosen auxiliary data during the finetuning phase on the target task. They show that using the appropriate source data during transfer learning improves the excess risk bound on the target task and propose a novel selection strategy to determine which subset of source data to use during finetuning, based on unbalanced optimal transport. A notable difference is that they choose source data in a class-wise manner, whereas we pick individual examples.

Overall, while previous work has designed data selection strategies to boost transfer, we consider the different problem formulation of transfer unlearning and an (adaptation of a) recent relaxed definition of unlearning that is valuable in practice. We initiate this line of research by empirically investigating the utility of simple selection mechanisms.

\section{Methodology}
\label{sec:method}

In this section, we begin by adapting existing methodology from the literature to transfer unlearning setting. We then introduce our proposed approach that uses data selection to replace non-static target examples and demonstrate that it is more efficient than alternatives while performing exact unlearning according to \cref{defn:unlearninggeneral-transfer}.

\subsection{Exact unlearning by retraining} \label{sec:exact_by_retraining}
As mentioned previously, exact unlearning under \cref{defn:dpunlearning}, and, similarly, under \cref{defn:dpunlearning-transfer}, in deep neural networks is possible only with (variants of) retraining the model from scratch excluding the data to be deleted. In the case of transfer unlearning, unlearning a set $S' \subseteq \forgetset$ by retraining (or, more precisely, ``retransferring'') amounts to running $\Alg(\ModelSource, S \setminus S')$, i.e. rerunning transfer learning excluding $S'$. 
This procedure, however, can be computationally expensive if reran for each deletion request.

A more efficient approach would be to produce $\Alg(\ModelSource, S^{*})$ once, where $S^{*} \subseteq \retainset \subset S$, and thus will not receive deletion requests. 
This variant is exact unlearning according to \cref{defn:unlearninggeneral-transfer} but not
\cref{defn:dpunlearning-transfer}.
In exchange, it is more efficient due to amortization: it can be thought of as addressing all potential future deletion requests ahead of time. We will later describe our method that is equally efficient as this variant (and also exact according to \cref{defn:unlearninggeneral-transfer}) but performs significantly better on various datasets.

\subsection{Approximate unlearning} \label{sec:approx}
We can similarly adapt known approximate unlearning methods to transfer unlearning. 
Specifically, we would first obtain the target model via transfer learning using all available training data, yielding a model $\ModelTarget = \Alg(\ModelSource, S)$.
Subsequently, to serve a deletion request for $S' \subseteq \forgetset$, the approximately unlearned target model would be produced by running $\Unlearn(\ModelTarget, S')$. While, generally, a single application of $\Unlearn$ can be more efficient than retraining, the need to re-run unlearning for each deletion request can lead to overall inefficiency while it is also reported that approximate unlearning methods can hurt model utility, e.g. test accuracy (which may be further amplified in such a sequence). In contrast, our proposed method is efficient due to amortization, i.e. handling all deletion requests ahead of time. We later show that it also yields stronger (target) test accuracy compared to two approximate unlearning baselines.

\subsection{Transfer unlearning by data selection} \label{sec:ourmethod}
We now describe our proposed approach to transfer unlearning that utilizes a static auxiliary dataset $\DataAux$ to replace non-static target data, $\forgetset$. Our approach is comprised of two steps: (1) Data selection, and then (2) Transfer learning.

{\bf Data selection.} Let $\DataTarget = \retainset \cup \forgetset$ as above.
We consider a selection mechanism $g^*$ that chooses
a set $\DataAux' = g^*(\ModelSource,\forgetset,\DataAux)$, 
satisfying 
$\DataAux' \subseteq \DataAux$.
That is, given a pretrained model $\ModelSource$, $g^*$ selects additional training data from the available pool of auxilliary data, $\DataAux$. Our approach will be then to use all static data $\retainset$, and this subset $\DataAux'$ of auxiliary static data to train (notably, no examples from $\forgetset$ are used for training).

We now describe how our selection mechanism works.
%
For each class label $c$ in $\forgetset$, we perform the following steps:
\begin{enumerate}
    \item Compute the similarity (using dot-product in the embedding space of $\ModelSource$) between each example in $\DataAux$ and each non-static example of class $c$.
    \item Compute the \emph{average} similarity between each example in $\DataAux$ and all non-static samples of class $c$, using step 1.
    \item We sort the examples in $\DataAux$ in decreasing order of their average similarity w.r.t class $c$ to obtain an ordering $(x_1, y_1), \dots (x_{|\DataAux|}, y_{|\DataAux|})$.
    \item Select the top $M$ examples from that ordering and create a set $S^c_{\mathrm{sel}} = \{(x_1, c), \dots (x_M, c)\}$ that contains the $M$ examples from $\DataAux$ that are most similar (on average) to non-static examples of class $c$, relabeled as class $c$. Note that this approach allows us to select examples whose original labels in $\DataAux$ were different; in fact, the label spaces of $\DataAux$ and $\DataTarget$ may be different.
\end{enumerate}

Finally, we obtain the overall selected set $\DataAux'$ by taking the union of $S^i_{\mathrm{sel}}$ over all class labels $i$ appearing in $\forgetset$.

\begin{figure*}[h]
    \centering
    \includegraphics[height=15em]{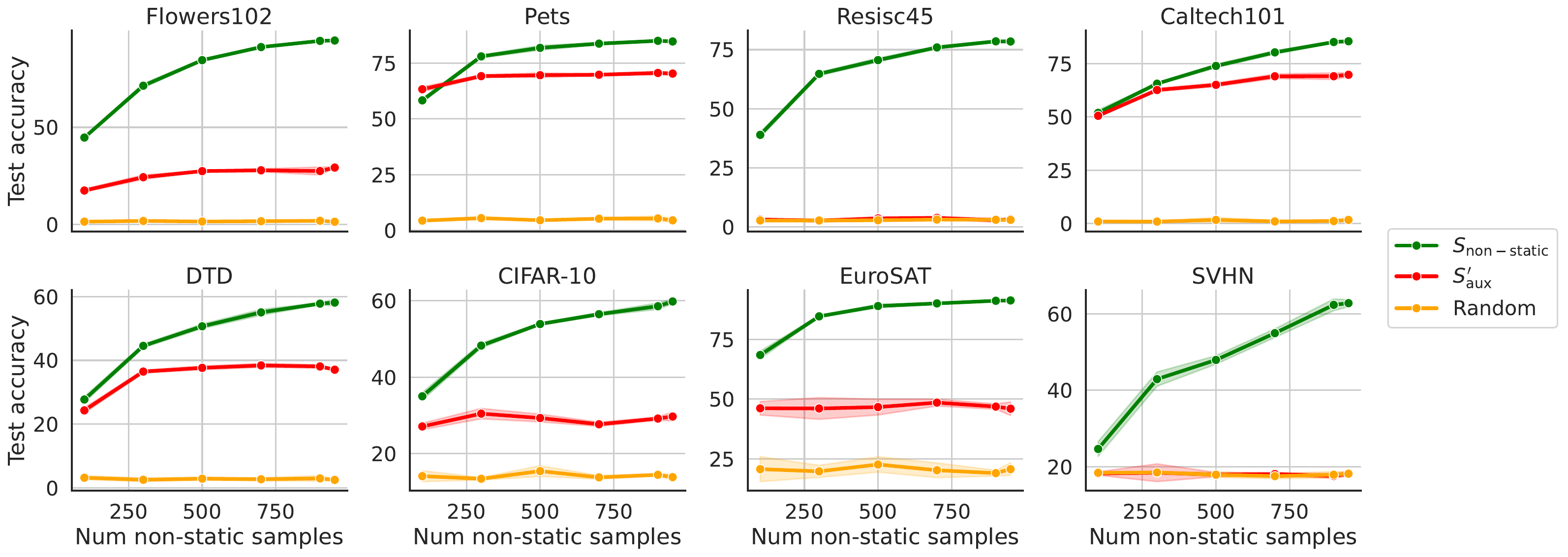}
    \caption{Target test accuracy for different $\forgetset$ sizes when $\retainset$ is empty ($\DataTarget = \forgetset$), obtained by finetuning $\ModelSource$ on three different training sets. \textbf{Our method (that finetunes only on $\boldsymbol{\selecteddata}$) in some cases approaches the upper bound ($\boldsymbol{\forgetset}$) of finetuning on all target data without unlearning} and greatly improves the control experiment of selecting random auxiliary examples (\textit{Random}).
    }
    \label{fig:Fig1}
\end{figure*}

{\bf Transfer by finetuning.} Given $\DataAux'$, we next perform finetuning on $\ModelSource$ based on the data $\retainset  \cup \DataAux'$.
We adopt this strategy due to its simplicity, effectiveness, and ubiquity in previous work and industry. More concretely, we first replace the task-specific output layer of $\ModelSource$ with a new randomly initialized output layer whose dimensionality is the number of classes in the target task. We then finetune the entire network using gradient descent on the target task. Since we consider classification tasks, we use the usual cross-entropy loss for this finetuning. 

{\bf A note on efficiency.} Due to never finetuning the model on any data from $\forgetset$, we effectively handle all unlearning requests ahead of time, similar to the amortized variant of retraining described in Section \ref{sec:exact_by_retraining}. However, we find that, our method yields substantially higher target test accuracy on many datasets compared to that approach.

Next, we demonstrate that our approach performs exact relative transfer unlearning, 
in the sense of \cref{defn:unlearninggeneral-transfer}.

\subsection{Data selection offers exact transfer unlearning}
\label{sec:unlearnwithdataselect}

\newcommand{\FTAlg}{\Alg_{ft}}
Consider a fixed $\DataAux$. Let $\FTAlg$ denote our finetuning with data selection algorithm, and let $S = \DataTarget \cup \DataAux =  \retainset \cup \forgetset \cup \DataAux$ be the available training data for transfer learning.
Our approach, described in \cref{sec:ourmethod}, maps $(\ModelSource,S) \mapsto \ModelTarget = \FTAlg(\ModelSource,\retainset\cup g^*(\ModelSource,\forgetset,\DataAux))$.
We now argue that this approach performs exact relative transfer unlearning for an identity unlearning algorithm $\Unlearn$.
 We show this by constructing a family $\{\Alg_{\theta}\}_{\theta \in \Theta}$ and map $\hat\theta$ corresponding to our data selection mechanism, such that $\Unlearn$ satisfies 
 \cref{defn:unlearninggeneral-transfer}.

Let $\Theta$ be the set of all subsets of $\DataAux$ and define  $\hat\theta(\ModelSource,S) = g^*(\ModelSource,\forgetset,\DataAux)$. 
For an arbitrary dataset of the form $S'' = \retainset \cup \forgetset' \cup \DataAux$, where $\forgetset' \subseteq \forgetset$, we define $\Alg_\theta(\ModelSource,S'') = \FTAlg(\ModelSource,\retainset \cup \theta)$. Finally, for every $S' \subseteq \forgetset$, we define
$\Unlearn(\ModelAny',S') = \ModelAny'$. 
To see that our transfer learning algorithm fits this definition, pick $S' \subseteq \forgetset$.
Then 
\begin{align*}
&\Alg_{\hat\theta(\ModelSource,S)}(\ModelSource, S \setminus S')
\\&=
\FTAlg(\ModelSource,\retainset \cup \hat\theta(\ModelSource,S))
\\&=
\FTAlg(\ModelSource,\retainset\cup g^*(\ModelSource,\forgetset,\DataAux))
\\&= \ModelTarget = 
\Unlearn(\ModelTarget, S').
\end{align*}
Intuitively, our method uses $\forgetset$ to inform data selection, but never finetunes on $\forgetset$ itself. That is, it uses  $\forgetset$ only for the purpose of instantiating $\theta^* = \hat \theta (\DataSource,S)$ to select a member of $\{\Alg_{\theta}\}_{\theta \in \Theta}$. Exact relative unlearning is therefore achieved for a trivial $\Unlearn$ since that instantiated algorithm $\Alg_{\theta^*}$ operates on a dataset that excludes $\forgetset$.

\begin{figure*}[h]
    \centering
    \includegraphics[height=15em]{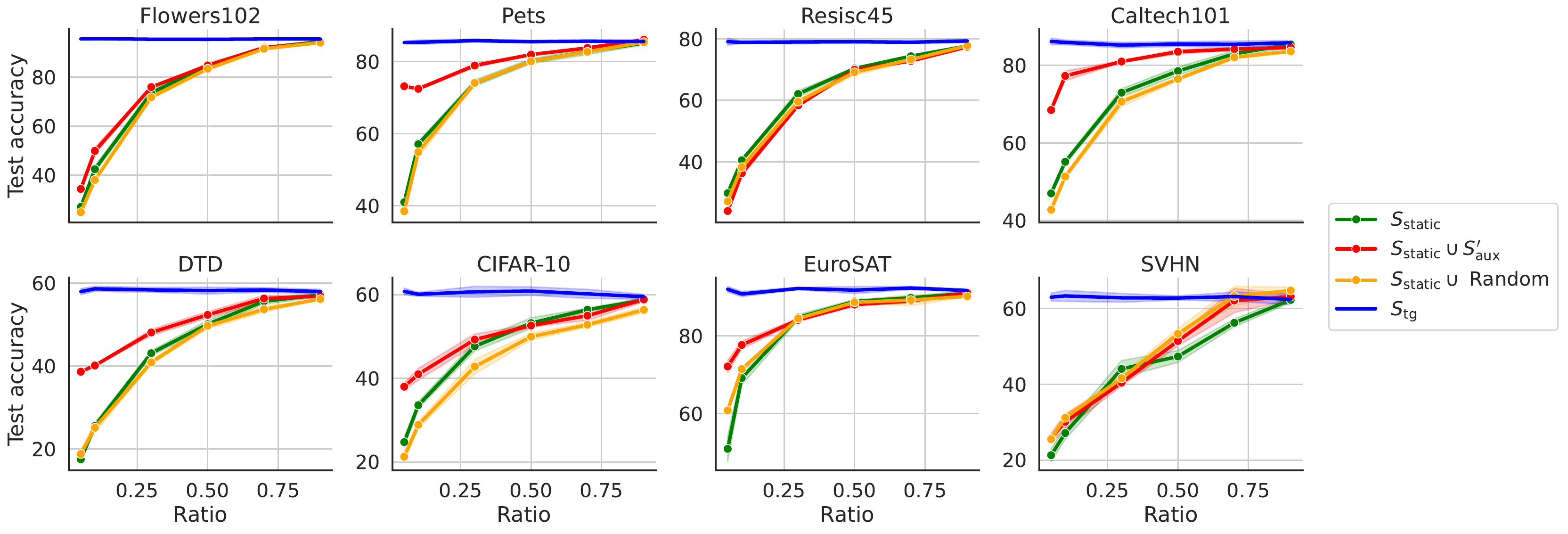}
    \caption{Target test accuracy for the scenario where $\retainset$ is non-empty (with different \textit{ratios}), obtained by finetuning $\ModelSource$ on four different training sets. \textbf{Our method (\boldsymbol{$\retainset \cup \selecteddata}$) in some cases approaches the upper bound ($\boldsymbol{\DataTarget}$) of finetuning on all target data without unlearning and greatly improves both the control experiment of selecting random auxiliary examples ($\boldsymbol{\retainset \cup $}  Random), and the exact unlearning baseline ($\boldsymbol{\retainset}$)}. While we do not boost the accuracy of $\retainset$ in every dataset, 
    the fact that our method outperforms the gold standard exact unlearning baseline in several cases (especially for small $\retainset$) is an important step forward.
    }
    \label{fig:Fig2}
\end{figure*}

\section{Experimental Investigation}
\label{sec:results}

In this section we investigate the effectiveness of our method by conducting a series of experiments across various datasets. We start by describing the experimental settings and then proceed to present the results.

\textbf{Datasets.} We consider ImageNet \cite{imagenet_cvpr09} as the source dataset for all our experiments; a common choice in the literature due to its size and diversity. We then consider nine diverse and widely used datasets in the field of computer vision as our target tasks: SVHN \citep{netzer2011reading}, CIFAR-10 \citep{krizhevsky2009learning}, EuroSAT \citep{alem2020deep}, Pets \citep{parkhi2012cats}, Flowers102 \citep{nilsback2008automated}, Caltech101 \citep{fei2006one}, DTD \citep{cimpoi2014describing}, Resisc45 \citep{cheng2017remote} and finally, the Retinophaty \footnote{The results presented in this section do not include the Retinopathy dataset; for a detailed explanation, please refer to \cref{retino}.} \citep{retinopathy} dataset. The datasets differ in terms of image quantity, class count, image dimensions and degree of perceived similarity to the source dataset. Additional details about each dataset can be found in \cref{datasets}. For each target dataset, only a subset of 1000 samples is used as the train set. The reason for this decision in explained in \cref{implement}. 

\textbf{Experimental setup.} We consider $\DataSource$ to play the role of $\DataAux$, which reflects a practical scenario where the source data is available and some subset of it may be relevant to the target task. For each target dataset, $\DataTarget$ is randomly partitioned into $\retainset$ and $\forgetset$ at varying ratios. Throughout this section, the term \textit{ratio} denotes the size of $\retainset$ relative to $\DataTarget$ with values of \{.05, .1, .3, .5, .7, .9\} (with lower ratio values corresponding to fewer examples in $\retainset$ and more examples in $\forgetset$). 
Further, we consider two distinct cases: an extreme case where $\retainset$ is empty, meaning that the \textit{ratio} is 0 and any example from $\DataTarget$ may need to be deleted, as well as a softer case where both $\retainset$ and $\forgetset$ are nonempty, where we consider different \textit{ratios}.

For simplicity and consistent with the most common setting in the literature \citep{golatkar2020eternal,kurmanji2023towards}, we study the case where we unlearn $S' = \forgetset$, rather than a strict subset $S' \subset \forgetset$. We note that, since our method handles the deletions of any $S' \subset \forgetset$ in an amortized manner anyway, it would not be different in a sequential scenario of unlearning different subsets $S'$. In contrast, approximate unlearning may suffer even more in that case. We leave that exploration for future work. 

\textbf{Implementation details.} We conduct hyperparameter tuning for the learning rate and the number $M$ of selected examples per target class.
The optimal learning rate value is determined for each dataset and ratio separately, considering values $\{0.001, .005, .01, .05, .1\}$. We additionally sweep over values for the total size of $\DataAux$, using either 10\% or 20\% of $\DataTarget$'s size.  Further implementation details are available in \cref{implement}.

\subsection{Experimental Results}
We now discuss our experiments, designed to evaluate the efficacy of our proposed method under different target tasks and scenarios, against relevant baselines: exact unlearning by finetuning $\ModelSource$ on only $\retainset$ (Section \ref{sec:exact_by_retraining}) and two approximate unlearning methods (Section \ref{sec:approx}). We present our investigation aimed at answering the below questions. 

\textbf{Q1: How well does our method perform when all target data is non-static?} In this scenario, we assume an empty static set (\textit{ratio} = 0). We compare the results of finetuning $\ModelSource$ on three different training sets: 1) $\forgetset$ (i.e. all of $\DataTarget$, since $\retainset$ is empty), 2) $\selecteddata$ obtained using our proposed mechanism $g^*$ and 3) \textit{Random}, a random subset of $\DataAux$ of the same size as $\selecteddata$.
Note that, for an empty $\retainset$, finetuning on $\selecteddata$ corresponds to our proposed method (it finetunes on $\retainset \cup \selecteddata = \selecteddata$). \textit{Random} acts as a control experiment where, for each target class $c$, $M$ examples are chosen at random from $\DataAux$ and then relabeled as class $c$. Finetuning on $\forgetset$, on the other hand, serves as an upper bound for the test accuracy obtained by directly finetuning on the non-static data without supporting unlearning. The results for are shown in \cref{fig:Fig1}.

We observe that, for the majority of the datasets, (1) our approach outperforms \textit{Random} by a large margin, suggesting that its success is not due to merely increasing the size of the training set and, (2) in certain cases, it bridges the gap with the upper bound, i.e. the oracle of finetuning on the non-static data without unlearning. The degree to which that gap is narrowed varies significantly among datasets, with some, such as Pets and Caltech102, getting very close; while others, like Flowers102 and EuroSAT, still exhibit a notable gap. We refer to datasest where data selection is successful as \textit{positive}. In contrast, for certain \textit{negative} datasets, using selected examples is not advantageous (SVHN and Resisc45). Note that changing the size of $\forgetset$ does not affect $M$; it only alters the number of per-class target examples used to compute the similarity metric. We also notice that achieving good performance with our method does not require a large number of non-static samples, and in fact increasing the non-static portion size eventually ceases to enhance its performance. Finally, we show examples of $\selecteddata$ for both positive and negative datasets in \cref{qualitative} and discuss later factors that affect the success of data selection. 

\textbf{Q2: How effective is our method when some of the target set is static?} We now study this softer case, where parts of the target dataset are static, under various \textit{ratios}. We consider finetuning on four different training sets this time: 1) $\retainset \cup \selecteddata$, corresponding to our approach, 2) $\retainset \cup$ Random, a control experiment as before, 3) $\retainset$, exact unlearning by training on only static data (Section \ref{sec:exact_by_retraining}) and 4) $\DataTarget$, acting as the upper bound for test accuracy obtained by finetuning on all target data without any unlearning.

\cref{fig:Fig2} depicts the test accuracy for various \textit{ratio} values across all datasets. We again notice that, for the positive datasets, our approach of adding $\selecteddata$ to $\retainset$ boosts its performance and reduces the gap with the the upper bound. The improvement is specially enhanced when the \textit{ratio} is smaller, i.e. when the non-static set constitutes the majority of the $\DataTarget$, and the static set forms a smaller portion. While we don't boost the accuracy of $\retainset$ in every dataset, our results represent important progress: that baseline is the gold standard exact unlearning method which we can outperform in several cases, especially for small $\retainset$.

\begin{figure}[h]
    \includegraphics[height=8.5em]{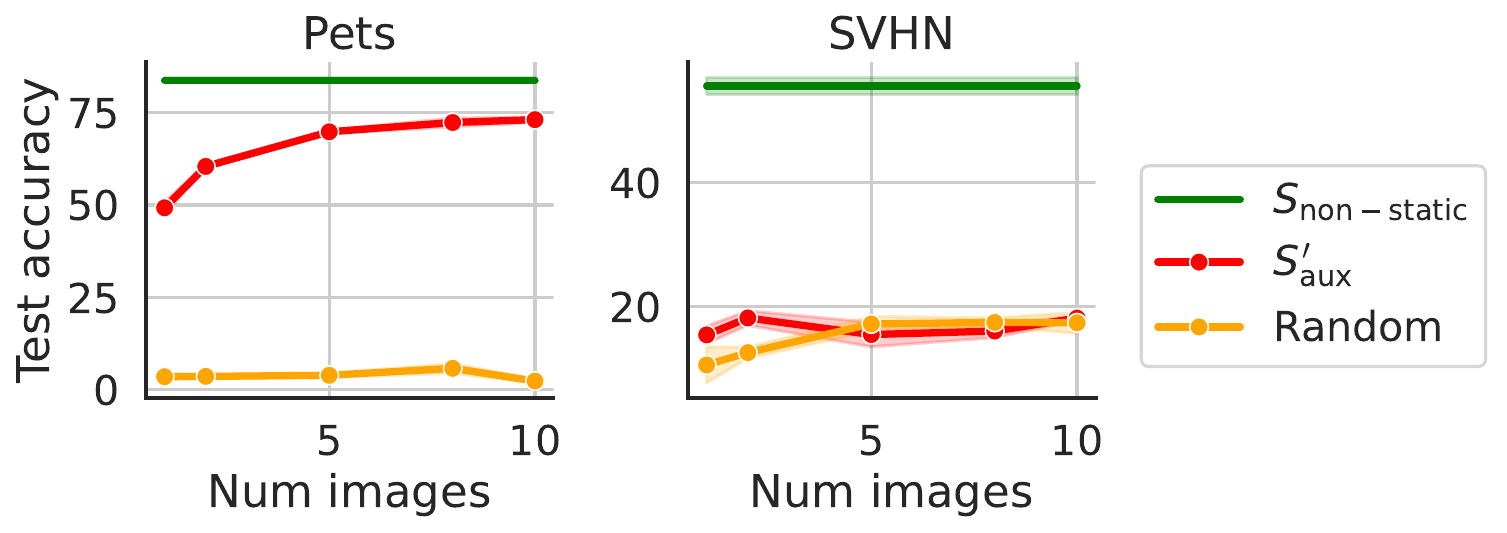}
    \includegraphics[height=8.5em]{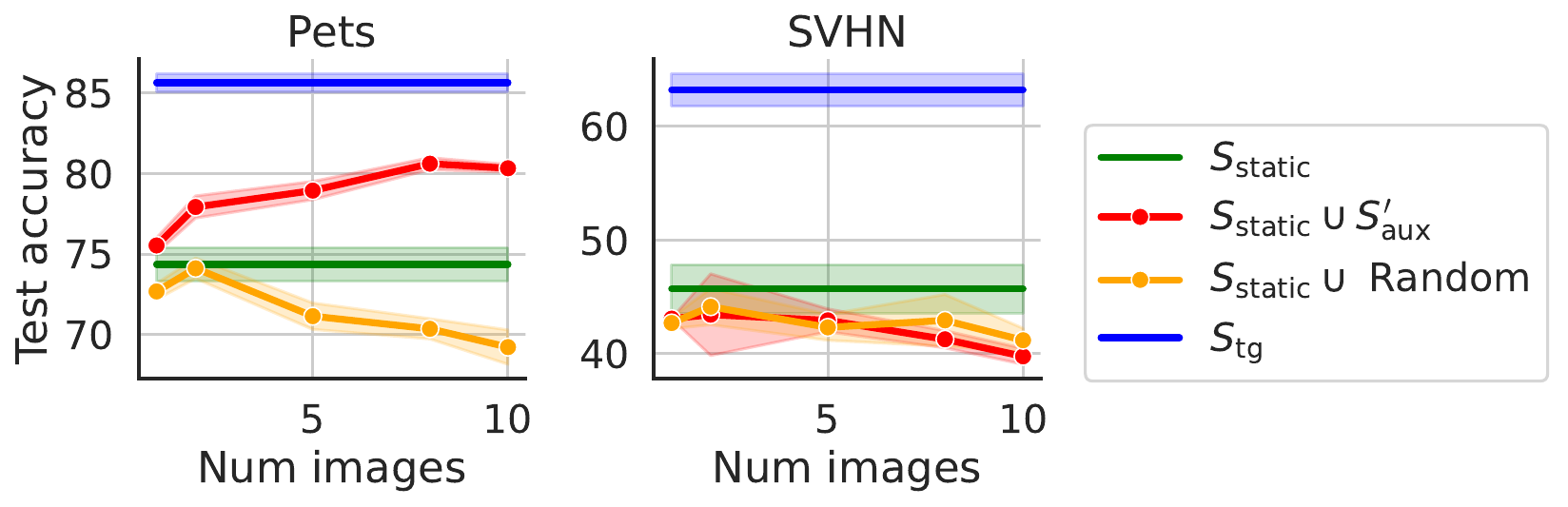}
    \caption{Target test accuracy for different number of selected examples for the case of empty $\retainset$ as in Q1 (left) and nonempty as in Q2 (right). As expected, increasing this number brings our method closer to the upper bound of finetuning on all target data without unlearning for the positive but not the negative dataset.  
    }
    \label{fig:Fig3}
\end{figure}

\textbf{Q3: What is the effect of the size of $\selecteddata$?} We now examine the impact of $M$ (the number of selected examples)
for a positive (Pets) and a negative dataset (SVHN). Here, we assume the \textit{ratio} stays the same while M varies. The results for $\textit{ratio} = 0.3$ is presented in \cref{fig:Fig3}, showing that, for the positive dataset, increasing the number of added similar images leads to improved performance for both empty and nonempty $\retainset$. Conversely, for the negative dataset, adding more source images either harms the performance (empty $\retainset$) or fails to provide any benefit (nonempty $\retainset$). This suggests that, for positive datasets, we could boost our results even further, by selecting a larger $\selecteddata$, whereas for the negative dataset, failure to improve results by adding $\selecteddata$ was not simply due to not adding enough examples; we examine other reasons below.   

\begin{figure}
\centering
\begin{subfigure}{0.4\columnwidth}
    \hspace{3em}
    \includegraphics[width=\columnwidth]{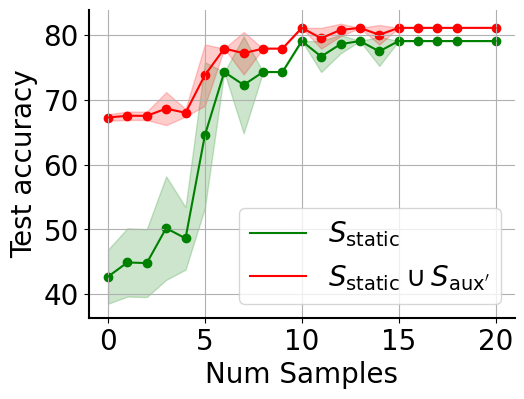}
\end{subfigure}
\hfill
\begin{subfigure}{0.4\columnwidth}
    \hspace{-3em}
    \includegraphics[width=\columnwidth]{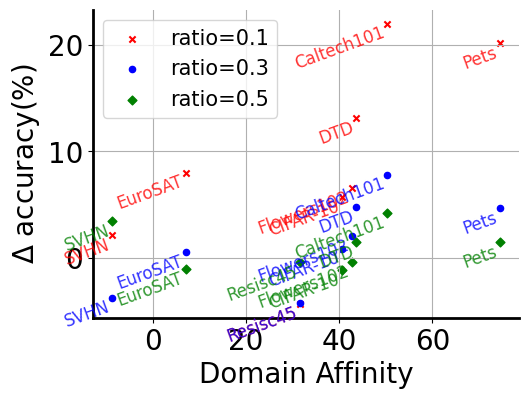}
\end{subfigure}
\caption{\textbf{Left}: Per-class target test accuracy for varying number of per-class samples in the $\retainset$ for the Pets dataset. Finetuning on only $\retainset$ is in green while finetuning on $\retainset \cup \selecteddata$ is in red. The \textit{ratio} for this set of experiments is fixed at 0.3. \textbf{Right}: The increase in test accuracy by finetuning on $\retainset \cup \selecteddata$ over finetuning on $\retainset$, as a function of Domain Affinity. Domain affinity correlates with the success of our approach.}
\label{fig:Fig4}
\end{figure}

\textbf{Q4: Which target classes benefit the most?} 
Depending on the \textit{ratio} value, the number of per-class samples in the $\retainset$ varies drastically, with small static sets potentially lacking representation from some classes. The addition of $\selecteddata$ is expected to be particularly beneficial in such instances. In \cref{fig:Fig4} (left), we plot the average per-class accuracy for various class sizes in Pets dataset's static set. The graph compares per-class accuracy when the network is finetuned on $\retainset$, versus $\retainset \cup S_{\mathrm{aux}}'$. As expected, classes with no or few samples in $\retainset$ show significant improvement with the incorporation of $\selecteddata$. However, even classes with more samples see performance gain, albeit less.    


\textbf{Q5: How does the relationship between $\DataAux$ and $\DataTarget$ affect the success of data selection?}
We hypothesize that the boost obtained by adding $\selecteddata$ depends on the relevance of $\DataAux$ to $\DataTarget$.
We quantify this notion using the \textit{Domain Affinity} score \citep{evci2022head2toe}, designed to measure the distributional shift between two domains. In \cref{fig:Fig4} (right), the performance gain achieved by adding $\selecteddata$ is plotted against the Domain Affinity score of all eight datasets, considering a few \textit{ratio} values. Our findings support the hypothesis that increased similarity between $\DataAux$ ($\DataSource$ in our case) to $\DataTarget$ correlates with greater boost by adding $\selecteddata$.

\textbf{Q6: How does our method compare to approximate unlearning methods?} We consider two widely used approximate learning baselines: \textit{Unlearn-NegGrad} and \textit{Unlearn-FineTune} \cite{golatkar2020eternal}. As explained in Section \ref{sec:approx}, both methods first finetune $\ModelSource$ on all $\DataTarget$, and then postprocess that model to unlearn the required forget set. As before, we consider a case where the forget set is all of $\forgetset$ (though in practical scenarios we could have a sequence of forget sets originating from  $\forgetset$). The methods we consider unlearn  $\forgetset$ either by gradient ascent on $\forgetset$ (\textit{Unlearn-NegGrad}) or gradient descent on $\retainset$ (\textit{Unlearn-FineTune}).
For these unlearning methods, we pick the checkpoint of unlearning that achieves the closest performance on $\forgetset$, as it would have if $\forgetset$ was not included in the training process (serving as the proxy for the ``best unlearning''). 
We include additional details in Appendix \ref{aprx}. From \cref{fig:Fig6}, we observe that our method performs no worse than approximate unlearning (and in fact, significantly better on the positive dataset), without ever finetuning on any part of $\forgetset$, and satisfies exact relative transfer unlearning, in contrast to those methods.


\begin{figure}[t]
    \centering
    \includegraphics[width=.99\columnwidth]{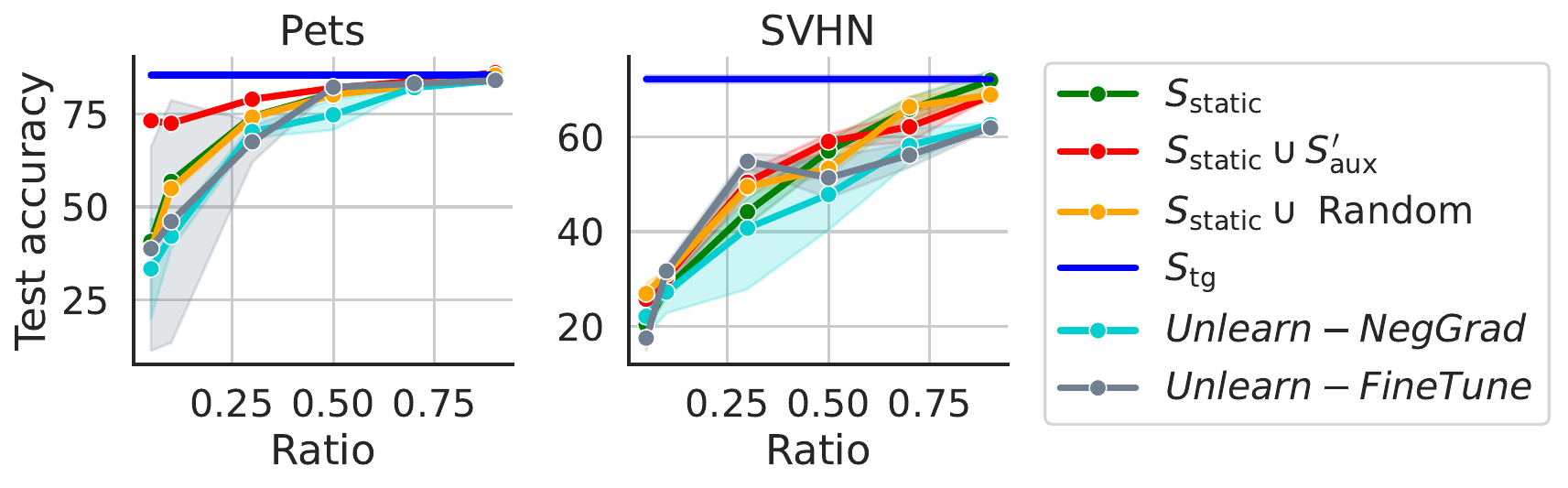}
    \caption{Target test accuracy for different \textit{ratios}, including comparison with two approximate unlearning methods: \textit{Unlearn-NegGrad} and \textit{Unlearn-FineTune}. We notice that our approach performs no worse than those methods, outperforming them by a large margin on the positive dataset, while being more amortized and satisfying exact relative transfer unlearning, unlike those methods. 
    }
    \label{fig:Fig6}
\end{figure}

\section{Discussion and Conclusion}
In this work, we have pioneered the use of data selection mechanisms for transfer unlearning 
for non-privacy applications. Specifically, this is the goal of transferring a pretrained model to target tasks in a way that efficiently supports unlearning requests for non-static target data whose permission of use has changed, disallowing their use for training purposes. We argue that privacy-oriented definitions of unlearning are too stringent for certain applications, and, to address this, we adapt a recent relaxed notion of unlearning to our problem formulation of transfer unlearning. We also demonstrate that our method that replaces non-static target data with selected auxiliary data is an exact unlearner under that definition, while being more efficient (highly amortized) than alternative methods.

While the problem formulation we consider is intentionally different from privacy, an interesting direction for future work is to study the privacy properties of our selection mechanism. Note, for instance, that the averaging operation that we perform can be seen as an addition of noise, and formally characterizing the level of privacy it affords is an interesting future study. Furthermore, mapping more than one non-static target examples to the same auxiliary example can contribute to protecting privacy: knowledge of the selected auxiliary examples may not be traced back to any particular non-static target example.

We emphasize as well that even exact unlearning approaches that satisfy stringent privacy-oriented definitions of unlearning have drawbacks that should be weighted differently depending on the application of interest. For example, \citet{chen2021machine} demonstrate that, even exact unlearning approaches can comprise privacy under some circumstances: they are vulnerable to attackers that have access to the model both before and after unlearning is performed. Instead, our method trivially defends this type of attack: since we never train on the non-static data, we alleviate the need to produce a post-processed model to address any unlearning request, effectively addressing all possible unlearning requests ahead of time.

Our empirical evaluation of our method across multiple datasets and against various baselines reveals promising results for this approach. For some datasets (especially when the static portion of the target set is small), we achieve target test accuracy significantly higher than finetuning on only that static set (i.e. exact unlearning by excluding the non-static set), in some cases even approaching the oracle reference point obtained by finetuning on all available target data without unlearning. We also outperform two common approximate unlearning baselines.

Our method's success hinges on the availability of a ``related'' (static) auxiliary dataset which may not always exist.
Nevertheless, in practical scenarios where this auxiliary data condition is met our approach may bring
 a substantial performance boost, as illustrated by our compelling experimental results comparing our method to the gold-standard exact unlearning baseline (retraining by excluding non-static set). 
Furthermore, in applications where auditing unlearning methods is difficult, maintaining exact unlearning by relaxing the definition, as done in our work, may be preferable to approximate approaches \citep{hayes2024inexact}. 
Such relaxed notions of unlearning may be particularly suitable and useful in scenarios where differential privacy guarantees are not the chief concern.
Our work encourages the community to consider relaxations of the common differential privacy-based notion of unlearning, as done in our work.
We argue that our adaptation of relative unlearning to transfer learning with data selection is an important step for many applications.


Unavoidably, we have merely scratched the surface of this new line of exploration; we discuss limitations further in Appendix \ref{limitations}. We hope that future work considers more complex transfer learning algorithms, data selection mechanisms, architectures and types of pretrained models.

\subsubsection*{Broader Impact Statement}
The unlearning problem is intrinsically linked with notions of consent (for one's data to be used to train ML systems) and legal compliance (e.g., GDPR). There is currently no universally-accepted and measurable definition of unlearning, and to our best knowledge the legal standard for determining compliance with a deletion request has not yet been determined in the context of machine unlearning. Our work contributes to this ongoing conversation by considering a new definition for unlearning and exploring ways to guarantee unlearning by construction under this definition.


\subsubsection*{Acknowledgments}

The authors would like to thank Amr Khalifa, Ioannis Mitliagkas, Hugo Larochelle, and Daniel M. Roy for feedback on various drafts.


\bibliography{main}
\bibliographystyle{tmlr}

\appendix
\section{Appendix}

\subsection{Datasets}
\label{datasets}
In this section, a brief description of all nine datasets used in this work is provided. 

 \begin{itemize}
    \item \textbf{CIFAR-10}: CIFAR-10 consists of 60,000 color images, each of size 32x32 pixels. The images are  categorized into 10 classes, encompassing common objects such as various types of vehicles and animals. The dataset is evenly divided into 50,000 training images and 10,000 testing images.
    \item \textbf{SVHN}: SVHN contains images of house numbers captured from Google Street View and is widely used for digit recognition tasks. The images are sized at 32x32 pixels and belong to one of the 10 total classes.
     \item \textbf{Flowers102}: This dataset includes 102 categories of various types of flowers, totaling over 8,000 images. Although the original images were larger, the image size has been adjusted to 256 x 256, which is better suited for our task. 
     \item \textbf{Caltech101}: The dataset comprises images from 101 object categories, each with a minimum of 40 images. Widely used for object recognition and image categorization tasks, the original image sizes vary but being modified to 256 x 256 for the sake of the current task.
     \item \textbf{DTD}: DTD features a diverse collection of 47 texture classes each comprising 120 images. The original image sizes range between 300x300 and 640x640 but have been resized to 256 x 256. 
     \item \textbf{Pets}: The Pets dataset consists of images of cats and dogs belonging to 37 different breeds. It is commonly used for fine-grained classification tasks.Similarly, the images are resized to 256 x 256 dimensions. 
     \item \textbf{EuroSAT}: EuroSATA is a specialized dataset created for the classification of 64 x 64 Sentinel-2 satellite images, categorizing them into 10 distinct land-use categories.  
     \item \textbf{Retinopathy}: This dataset includes retinal images intended for the categorization of diabetic retinopathy severity, ranging form level 0, indicating a healthy state, to level 4, denoting the most severe conditions. Images are adjusted to a size of 256 x 256 pixels.
     \item \textbf{Resisc45}: The dataset comprises remote sensing images with a resolution of 256 x 256 pixels, representing 45 different scene categories and is designed for scenes classification in remote sensing applications.
 \end{itemize} 
 
\subsection{Implementation Details}
\label{implement}
We use the PyTorch \cite{pytorch}
framework for the implementation. Each image is normalized to have a zero mean and unit variance. The normalization step is preceded by data augmentation, where random cropping and horizontal filling are applied. A pre-trained ResNet-18 \cite{he2016deep} is finetuned for 100 epochs, with a weight decay value of 0.0005, using stochastic gradient with momentum ($\beta=.9$). During training, the learning rate value is dropped by an order of magnitude at epoch 80. The performance of the trained network is evaluated by its accuracy on an unseen test set. The results are reported for a checkpoint with the highest accuracy on a validation set held separately from the target train set used for training.All reported results are the average over 3 different runs. 

Finally, for each dataset, only a subset of 1000 training samples is used as the train set. This is primarily for practical reasons, ensuring inferior performance for the network trained on $\retainset$ compared to training on the entire $\DataTarget$. Specifically, if training on $\retainset$ is sufficient for achieving the same target test accuracy as training in all of $\DataTarget$, unlearning is trivial, so we choose to focus on settings that are more challenging. That said, we do vary the static to non-static  ratio in our experiments to study these effects in a soft manner.

\subsection{Retinopathy Dataset Results}
\label{retino}
\begin{figure}
    \centering
    \includegraphics[width=.6\columnwidth]{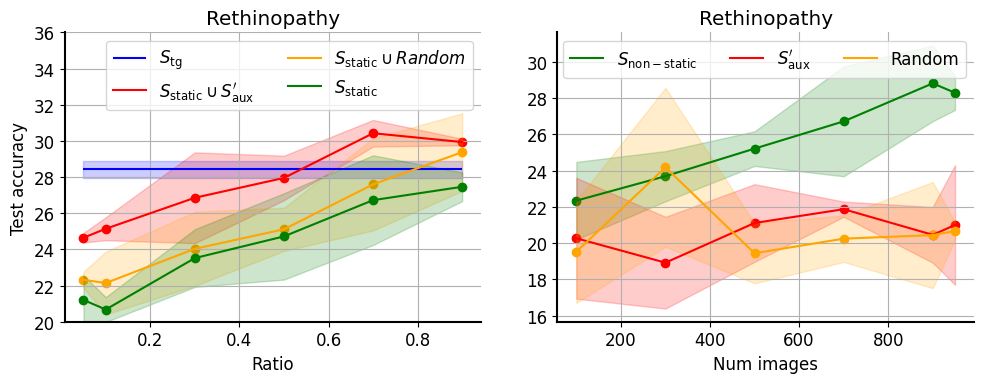}
    \caption{Balanced test accuracy is displayed for the case of an empty $\retainset$ in Q1 (left) and a non-empty $\retainset$ in Q2 (right). For the empty $\retainset$ scenario, the results are reported for various \textit{ratios}, whereas for the non-empty $\retainset$ case, the results are reported for different $\forgetset$ sizes. The \textit{ratio} value here is 0.}
    \label{fig:App2}
\end{figure}

Among the nine datasets selected to assess our method, Retinopathy is the only one exhibiting severe class imbalance, with around 70\% of samples belonging to a single class. Examining the effect of introducing source images to target classes becomes intricate in the presence of such class dominance, exceeding the score of this paper. Nevertheless, we report the \textit{balanced} test accuracy results for this dataset in \cref{fig:App2}.

We exclude these results from the main paper due to the technical difficulties caused by severe class imbalance which may require more sophisticated transfer learning approaches (e.g. using importance weighting); which is a difficulty and problem setting orthogonal to the one we set out to study in this work. While we observe encouraging results for our method here too, the variance is high, and the severe class imbalance may introduces confounding effects. We leave it to future work to study transfer unlearning for very imbalanced target tasks.

\subsection{Qualitative Results}
\label{qualitative}
\begin{figure*}
    \centering
    \begin{subfigure}{.8\textwidth}
        \includegraphics[width=\columnwidth]{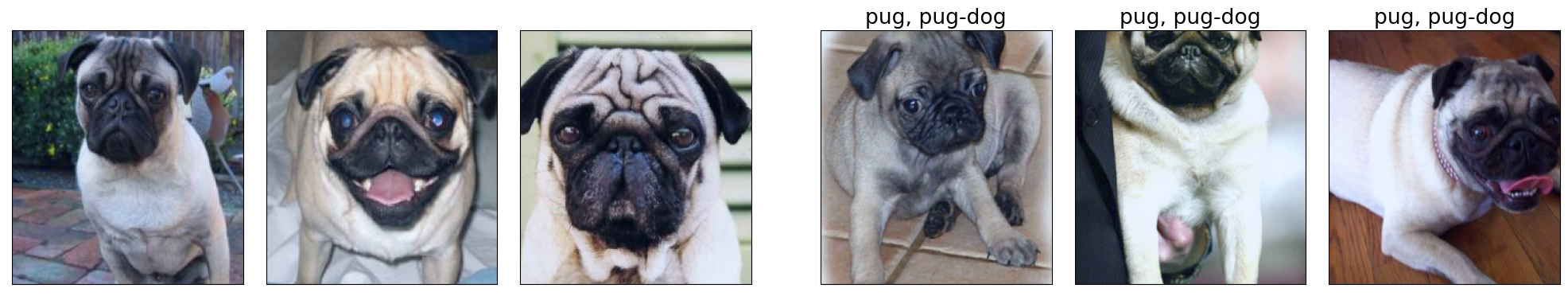}
    \end{subfigure}
    \begin{subfigure}{.8\textwidth}
        \includegraphics[width=\linewidth]{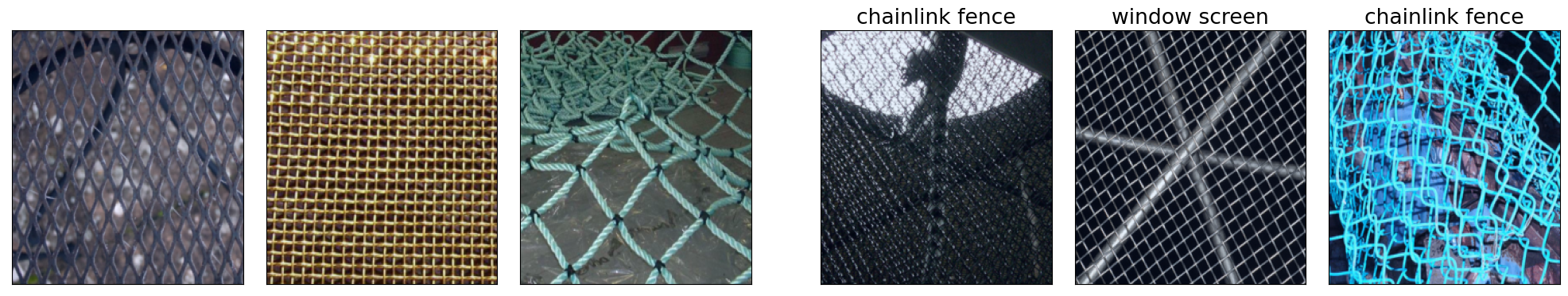}
    \end{subfigure}
    \begin{subfigure}{.8\textwidth}
        \includegraphics[width=\linewidth]{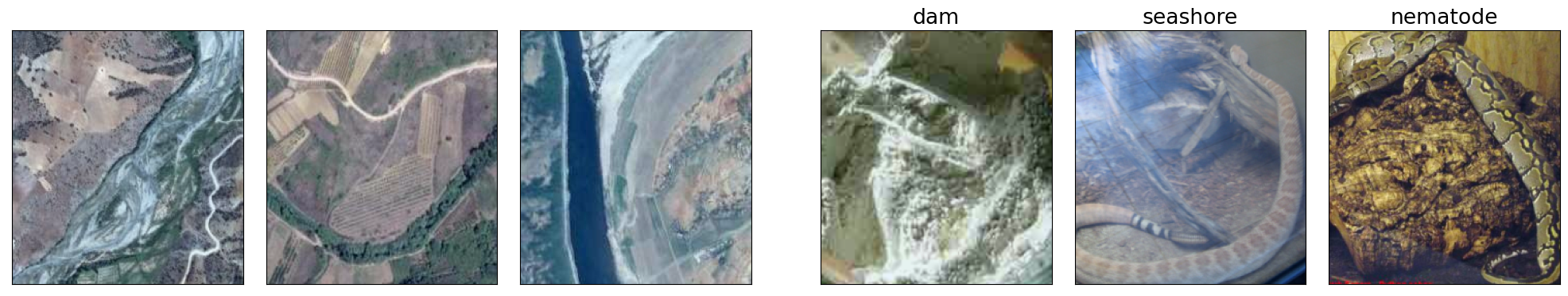}
    \end{subfigure}
    \caption{Three images from a non-static class set (left panel) are paired with three selected source images for the same class (right panel) across three dataset: Pets (top row), DTD (middle row) and Resisc45 (bottom row). The title of each selected source image displays its corresponding source class label.}
    \label{fig:App1}
\end{figure*}

To exemplify the images chosen through our data selection method, we have visualized three randomly chosen images from the most similar source images for a randomly chosen target class. These images are then juxtaposed with three randomly selected non-static images from the same class. The results for three datasets are shown in \cref{fig:App1}. 

\subsection{Approximate Unlearning}
\label{aprx}
To claim a successful unlearning method, it is crucial to demonstrate (1) effective unlearning of the non-static samples, and (2) no significant performance drop on the held-out target set.
In the case of exact unlearning, (1) is guaranteed (according to different notions), but for approximate unlearning methods, it is not. Therefore, we need to take care in comparing approximate methods against exact ones: simply reporting the results for the checkpoint with the highest validation accuracy is not appropriate. This is because, at one extreme (e.g. very small learning rate or number of steps), \textit{Unlearn-FineTune} and \textit{Unlearn-NegGrad} may have high target accuracy but at the expense of not having actually unlearned. Therefore, it is only fair to consider instantiations (e.g. hyperparameters) of approximate unlearning methods that demonstrate some acceptable level of unlearning, for a fair comparison with exact unlearning methods in terms of their test accuracy.

\begin{figure*}
    \centering
    \begin{subfigure}{.8\textwidth}
        \includegraphics[width=\columnwidth]{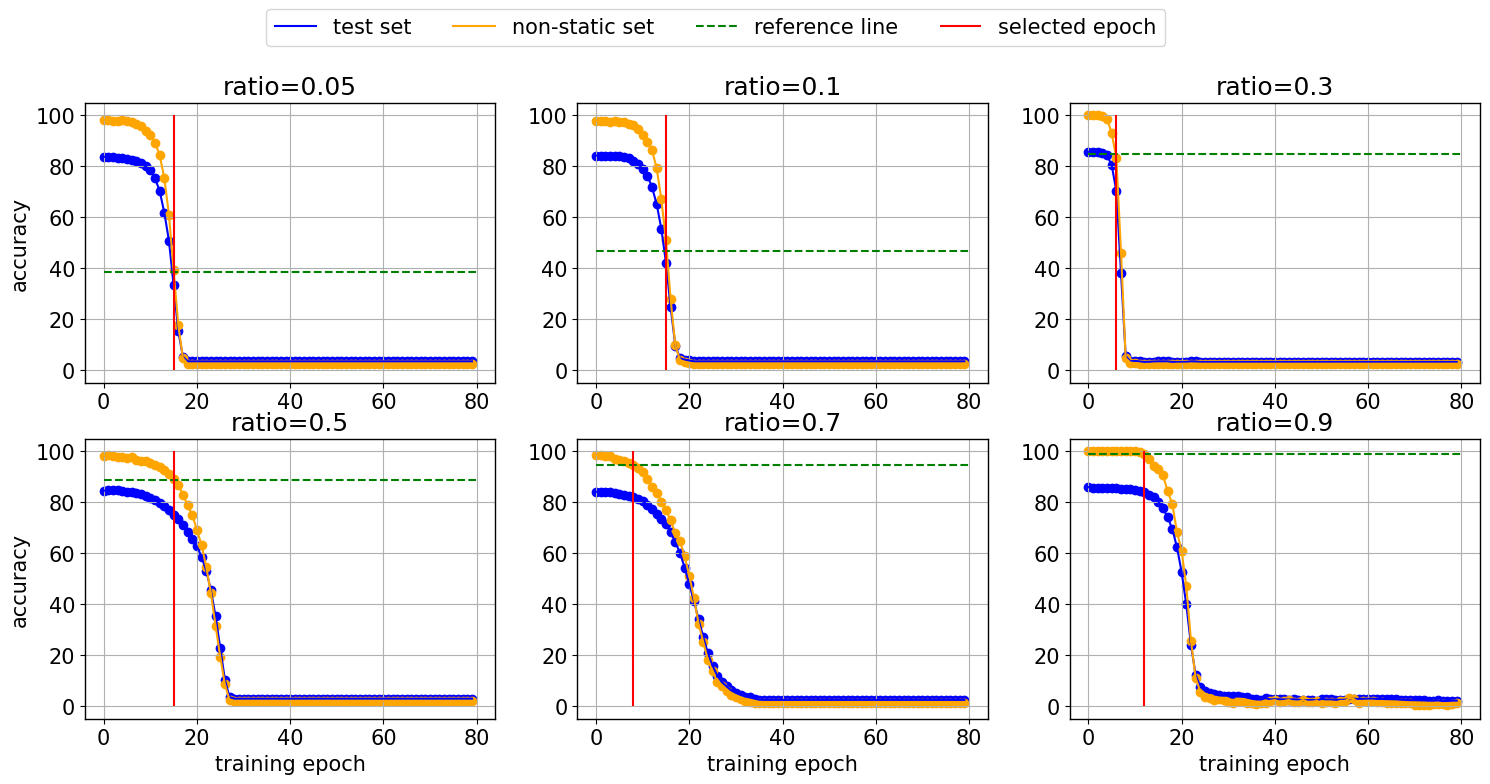}
    \end{subfigure}
    \begin{subfigure}{.8\textwidth}
        \includegraphics[width=\linewidth]{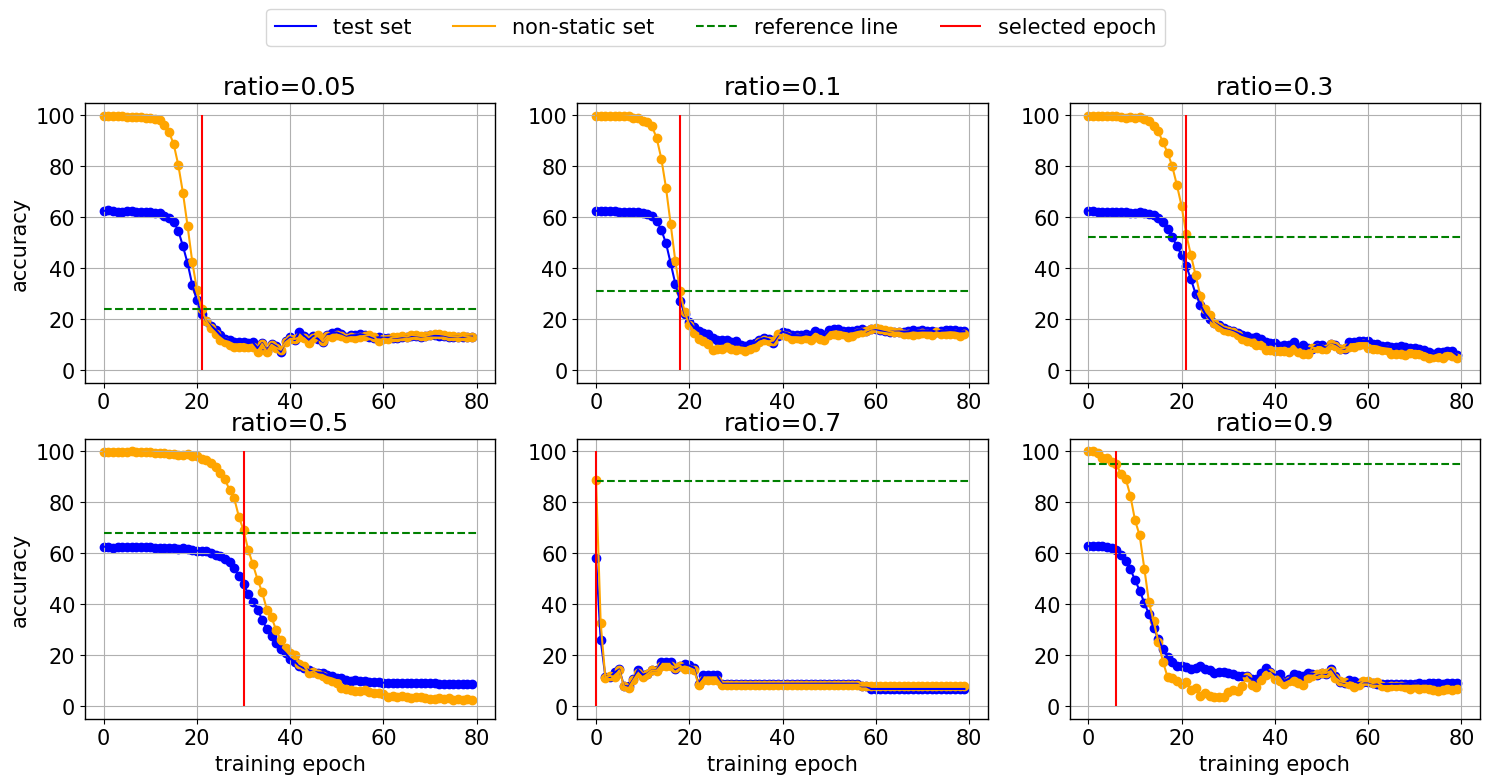}
    \end{subfigure}
    \caption{Network performance on non-static (orange) and test (blue) sets during unlearning conducted by \textit{Unlearn-NegGrad} for Pets (top) and SVHN (bottom) datasets is presented for multiple \textit{ratio} values. The green line in each plot depicts the reference point, while the red line represents the selected checkpoint (epoch).}
    \label{fig:App3}
\end{figure*}

\begin{figure*}[h]
    \centering
    \begin{subfigure}{.8\textwidth}
        \includegraphics[width=\columnwidth]{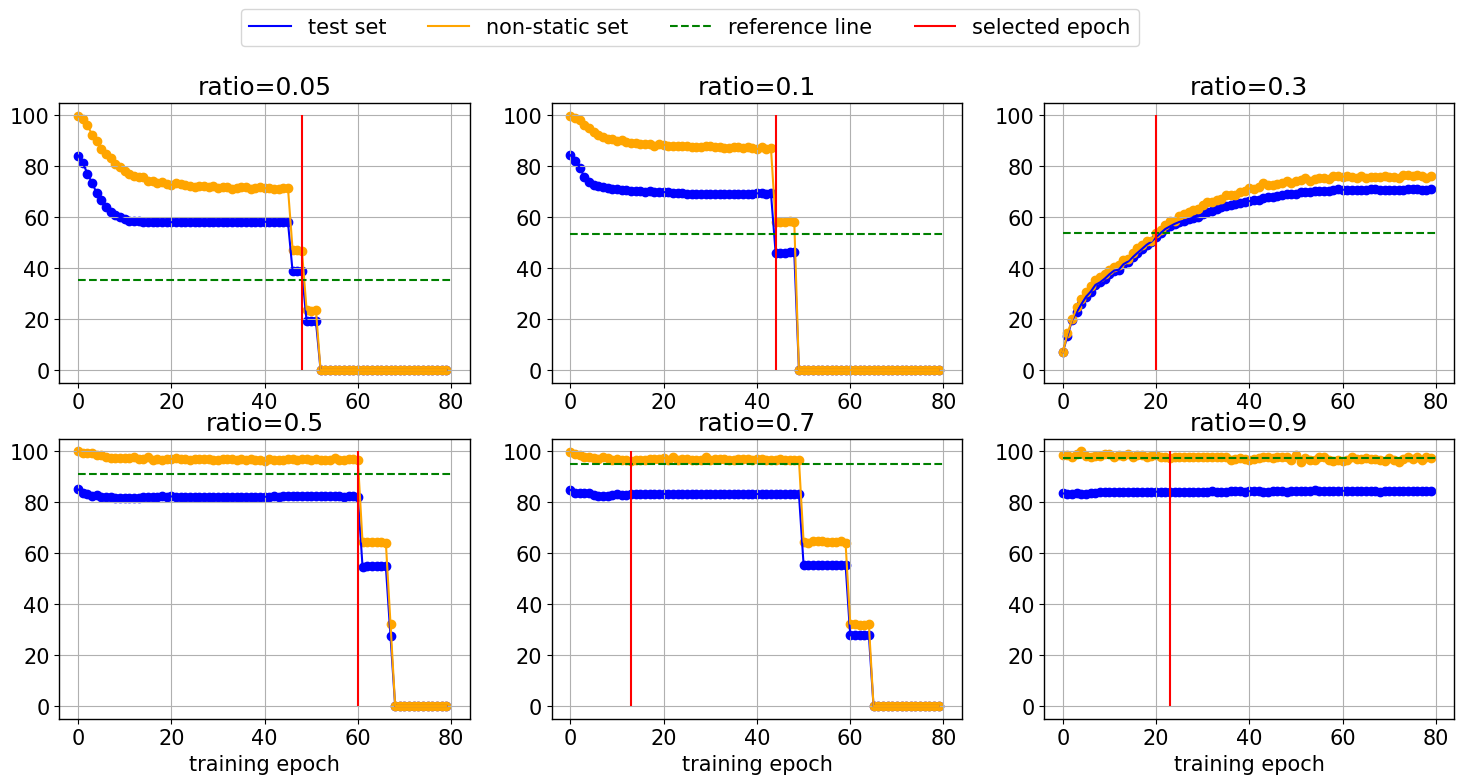}
    \end{subfigure}
    \begin{subfigure}{.8\textwidth}
        \includegraphics[width=\linewidth]{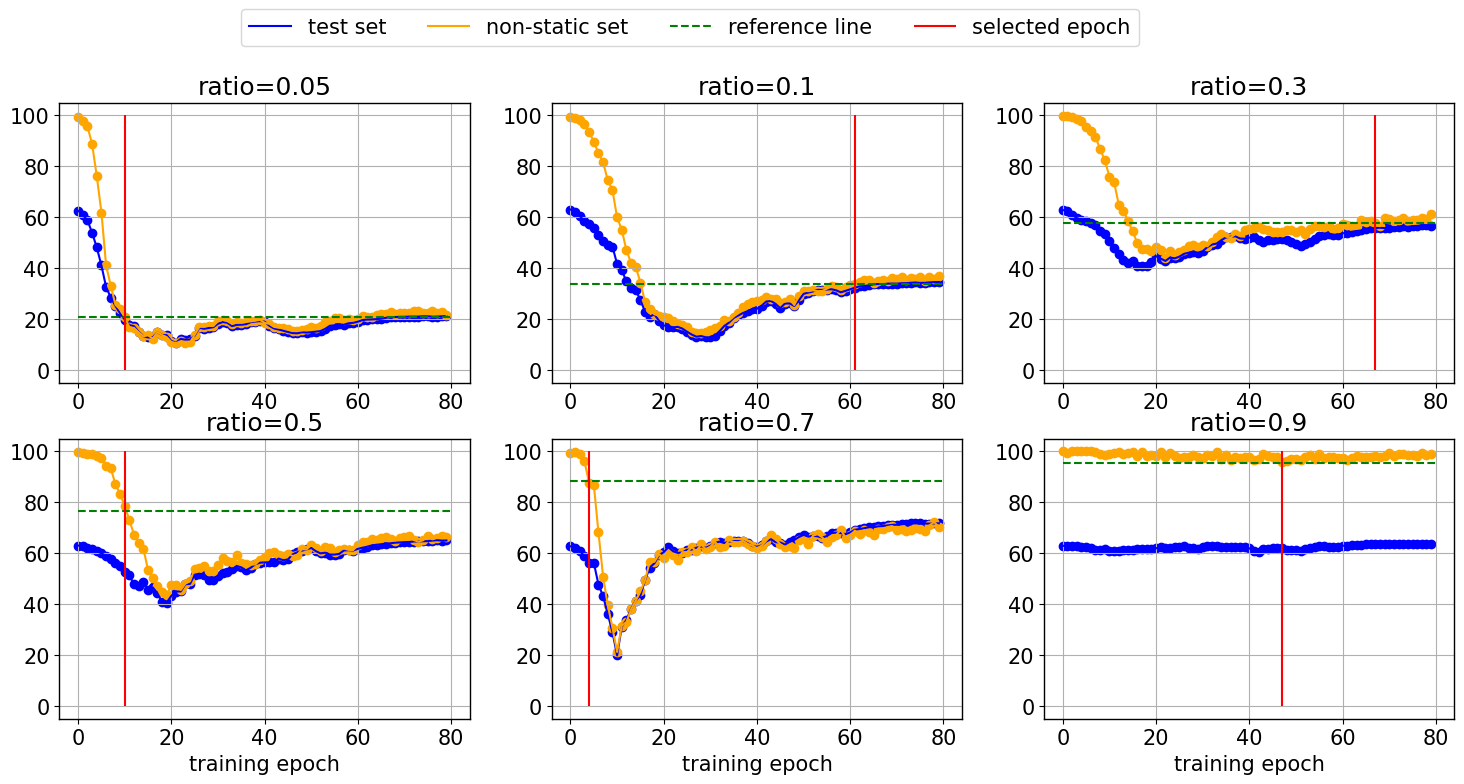}
    \end{subfigure}
    \caption{Network performance on non-static (orange) and test (blue) sets during unlearning conducted by \textit{Unlearn-Finetune} for Pets (top) and SVHN (bottom) datasets is presented for multiple \textit{ratio} values. The green line in each plot depicts the reference point, while the red line represents the selected checkpoint (epoch).}
    \label{fig:App4}
\end{figure*}

To address this issue, we use a simple proxy for ``good unlearning'' in order to pick the checkpoint of each approximate unlearning algorithm that we will use to compare against exact unlearning methods. Specifically, we use the $\forgetset$ accuracy of an identical network finetuned exclusively on the static set ($\retainset$) as the reference point for how high the $\forgetset$ accuracy should be, for ``good unlearning''.  We then select the checkpoint where the unlearned model's accuracy on $\forgetset$ is closest to the reference point and use it to report the test accuracy. This process is applied for both \textit{Unlearning-NegGrad} and \textit{Unlearning-FineTune}. The network performance on both the non-static and test sets, as well as the reference point and selected checkpoint, for \textit{Unlearn-NegGrad} and \textit{Unlearn-FineTune}, are displayed in \cref{fig:App3} and \cref{fig:App4}. Results are reported for Pets and SVHN datasets. It is important to note that the test accuracies at selected epochs (red vertical line) are the values reported in \cref{fig:Fig6}.   

\subsection{Limitations}
\label{limitations}

Importantly, a limitation of our approach is its reliance on the availability of an auxiliary dataset that is ``sufficiently similar'' to the target dataset. Specifically, we found that the degree of ``affinity'' between the source dataset (our auxiliary data) and the target dataset is strongly correlated with the boost in test accuracy observed by incorporating $\selecteddata$. Luckily, there are several scenarios with available public and diverse datasets where we expect our method to be valuable. However, we do not anticipate this approach to be universally applicable to every transfer task. 

Another limitation of this method is that, in this paper, the similarities are computed from representations learned by a network pre-trained on an upstream task within the source domain in a supervised fashion. Therefore, the representations learned in the penultimate layer are tailored specifically for the task at hand. This could potentially hinder the quality of $\selecteddata$, particularly in cases where the downstream task is significantly diverges from the upstream one. An apparent solution is to pre-train the network using unsupervised learning instead, which could be a direction for future work. We believe that this line of investigation can only further boost our results.  

\end{document}